# A Logic and Adaptive Approach for Efficient Diagnosis Systems using CBR

Ibrahim El Bitar
SIweb Team, SIR Laboratory
Ecole Mohammadia d'Ingénieurs, Mohammed Vth University-Agdal, Rabat, Morocco

Fatima-Zahra Belouadha
SIweb Team, SIR Laboratory
Ecole Mohammadia d'Ingénieurs, Mohammed Vth University-Agdal, Rabat, Morocco

Ounsa Roudiès
SIweb Team, SIR Laboratory
Ecole Mohammadia d'Ingénieurs, Mohammed Vth University-Agdal, Rabat, Morocco

## ABSTRACT
Case Based Reasoning (CBR) is an intelligent way of thinking based on experience and capitalization of already solved cases (source cases) to find a solution to a new problem (target case). Retrieval phase consists on identifying source cases that are similar to the target case. This phase may lead to erroneous results if the existing knowledge imperfections are not taken into account. This work presents a novel solution based on Fuzzy logic techniques and adaptation measures which aggregate weighted similarities to improve the retrieval results. To confirm the efficiency of our solution, we have applied it to the industrial diagnosis domain. The obtained results are more efficient results than those obtained by applying typical measures.

## 1. INTRODUCTION
The CBR (Case Based Reasoning) is a paradigm of intelligent reasoning which consists on reusing results of previously solved problems (source cases) to solve new problems (target cases) [2]. According to David Leake [7], the CBR is based on two assumptions about the nature of the real world: the regularity of situations, or similar problems, which must involve similar solutions, and the *recurrence* which assumes that future situations will probably be variations of current situations. In this context, the standard approaches of CBR propose a similarity-based mapping between the current problem (the target case) and the already solved problems (the source cases). The solution of the matched case will be proposed to solve the current target problem, as indicated in the analogy reasoning [6].

Typically, a case contains two parts: a situation description representing a problem and a solution used to remedy this situation. As showed in "Figure 1", the cyclic process of CBR includes five stages: Elaboration, Retrieve, Reuse, Revise and Retain all gravitating around a knowledge base (Case Base) of the application domain. [16]. We are interested in this paper to the retrieval phase using the available knowledge on the cases to find and select source cases that are similar to the target case.

Indeed, the retrieval phase is the most important point in the CBR process. It is based on similarity measures to retrieve source cases that are similar to the target case. However, several problems may arise during the extraction of similar cases. In order to achieve reliable results, this phase should address the problems of knowledge imperfections. In this scope, we propose a solution based on fuzzy logic techniques and adaptation measures aggregating weighted similarities. To evaluate our solution, we have applied it to the industrial diagnosis domain. Preliminary experiments showed that the results obtained, using the proposed solution, were efficient and more precise than those obtained by applying the measures proposed in the literature.

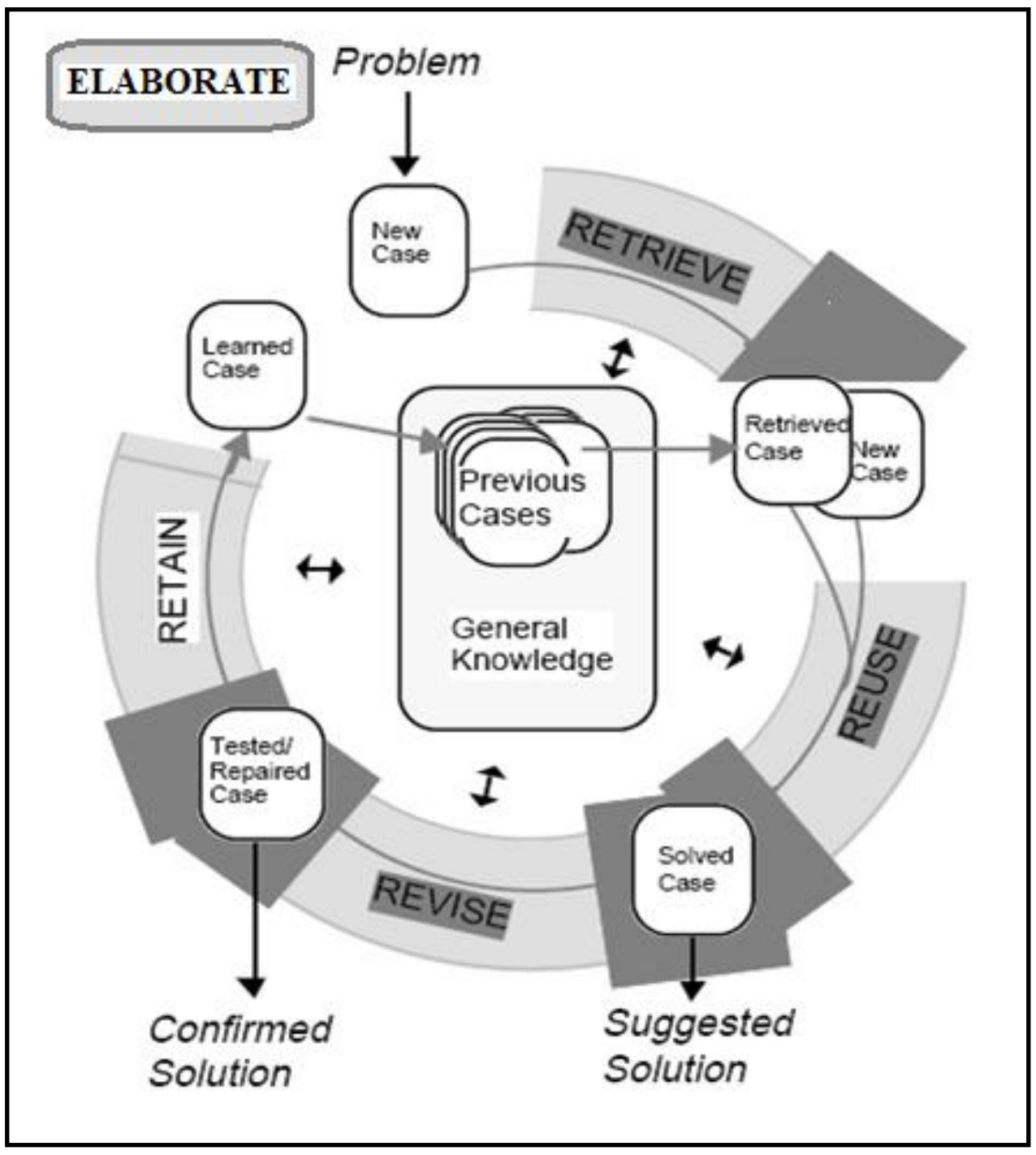

**Fig 1: CBR reasoning cycle**

The remainder of this paper is structured in seven sections. Section 2 presents the industrial diagnosis systems and the related existing systems. Section 3 describes the motivation of this work which aims to resolve the problem of knowledge imperfections. Section 4 presents the different types of these imperfections. The proposed solution is then described in section 5. A case study and the obtained results are presented and discussed in section 6. Section 7 concludes our work and identifies opportunities for reflection.

## 2. INDUSTRIAL DIAGNOSIS SYSTEMS
Several CBR systems and applications have been built in recent years to resolve various types of problems (diagnosis, planning, design, help desk, etc.). They have been used in different domains (medicine, engineering, law, etc.). However, CBR has been intensively deployed in the industry. It has been especially used to develop a lot of troubleshooting diagnosis systems dedicated to technical equipment. The purpose of this type of systems is to establish a diagnosis by identifying the failure component, and to propose an appropriate remedial action.

In this paper, we are interesting to the use of CBR for the industrial diagnosis systems. In the literature, various definitions of diagnosis systems have been given. Peng and Reggia [1], inspiring from the diagnosis etymology, consider a diagnosis system as a system which must analyze the observed events (symptoms, findings, etc.) that occur when running a given system. It must explain their presence, trace the causes, using knowledge about the system under consideration. Within this scope, the main objective of the reasoning is to understand the causes of the non-expected events when running a given industrial system [4], [10] and [13]. In the other terms, a diagnosis system is a system which is able to identify system failures, identify and locate causes of these failures, and proposes the actions to resolve them [11]. Many systems have been developed for different ends, such as, CaseLine, Checkmate, IRACUS, CASSIOPEE and FormTool. [3][5][8][12][14].

Checkmate is a conversational system which was proposed for industrial printers by Domino UK Ltd [5]. CaseLine is a help desk system that constitutes a demonstrator used by British Airways for the troubleshoot diagnosis [3]. CASSIOPEE is a system which which deals with the repair of Boeing 737 CFM aircraft engines [12]. IRACUS is dedicated to troubleshoot diagnosis for the locomotives used by the systems GE Transportation [8]. FormTool was developed to define and produce the plastic with different colors [14].

# 3. MOTIVATING SCENARIO

One of the key points of CBR is its retrieval phase. This phase computes similarity measures between the target case and the source cases. It consists in seeking correspondence between the cases descriptors. The local similarities are first held between descriptors of the same type and then aggregated to calculate the global similarity between two cases. In this context, several initiatives have been made towards the conception of cases similarity measures. The aim was to increase the accuracy of the mechanisms used to this end. In the literature, various and efficient mathematical techniques have been proposed. They enable discovering the source cases that are most similar to a target case.

However, all of these techniques assume that the knowledge brought by the descriptors of the target problem is reliable. This hypothesis poses a problem of results reliability when some knowledge provided through the descriptors is imperfect: rough, not well-defined, vague, incomplete or expressed in human language. The obtained results may indeed be erroneous.

The provided information may, in fact, be incomplete, uncertain or unquantifiable, and thereby leads to a knowledge imperfection. This imperfection is often due to several reasons which are, especially, related to errors in the capture of knowledge by humans or instruments that are generally subject to inaccuracies and uncertainties, the loss of information during the knowledge representation and theoretical assumptions that are based on flexible knowledge.

The presented work seeks to address the knowledge imperfection problems. It aims to propose a solution that should guarantee the reliability of the retrieval phase's results and ensure that the retrieved source cases are the most appropriate, even though knowledge imperfections exist.

# 4. KNOWLEDGE IMPERFECTIONS TYPES

As mentioned before, the knowledge provided on the target problems may be imperfect. These imperfections are of different natures. Bouchon-Meunier and Christopher have distinguished three types: imprecision, uncertainty and incompleteness [13].

The imprecision concerns digital knowledge as in the case of measurement errors (weight with 1% margin for example) or flexible knowledge (load capacity of a lift up to 4 or 5 persons).

The uncertainty reflects a doubt about the knowledge validity. It can, for example, be due to a relative reliability of the means used during the knowledge capture. It can also occurs when the observer intentionally give erroneous, incorrect or inaccurate information. Finally, a difficulty in knowledge obtaining or verification and forecasts are also examples of cases that lead to uncertainties.

As for the incompleteness, it constitutes a total or partial absence of knowledge. It is, in general, due to the inability to obtain certain information (e.g. forms which are not completely filled.) or to a problem occurring at the time of the knowledge capture (e.g. image with a hidden part).

These imperfections are not mutually exclusive and may occur at one time. Ignoring them, as is the case in most existing works, can lead to erroneous or non accurate retrieval results. Bouchon-Meunier and Christopher argue that the most satisfactory solution lies in the preservation of imperfections to a certain level, which makes it possible to not lose valuable information but also not give invalid results [13]. It is this balance between preservation of imperfection and simple treatment of the knowledge that our solution aims to achieve.

# 5. PROPOSAL SOLUTION

As we previously mentioned, the retrieval phase of CBR cycle is based on computing global similarities between the target case and the sources cases. We also remind that the global similarity measure is an aggregation of local similarities between the cases descriptors. However, the imperfection of knowledge detected in different descriptors can lead to erroneous measures.

Whatever the nature of the imperfection is and the reason behind, we have two possibilities. Either we ignore the imperfections and use an "accurate" representation that eliminates them, in other term we apply normalization through a standard defined by the system designer, or we keep them to exploit the information they contain and try to functionally represent them. Bouchon-Meunier says that the most satisfactory solution lies in imperfections conservation to a certain point, which consents to not lose valuable information and also to not give erroneous results [13]. We look for equilibrium between preservation of imperfection and simple treatment of the knowledge. In this vision, we propose our solutions to the imperfections in order to increase the accuracy of the search mechanism therefore improving the retrieval phase.

In this context, our solution aims to avoid invalid measures and to improve the retrieval results even if knowledge imperfections are noted.

To address the problem of imprecision and avoid the fact that it leads to conclude dissimilarity, we propose to use the techniques of fuzzy logic, in particular, the theory of possibilities that can correct imprecise data by means of mathematical formulas [19][20]. Reusing the proposed formulas at this stage allows us to approximate the values of imprecise descriptors in the target case to those of their corresponding descriptors in the source cases.

As for the case of uncertainty, we propose to eliminate the descriptors whose values are uncertain when calculating the similarity (i.e. no local similarity measure will be applied on these descriptors). Thereby, the global similarity will be obtained by aggregating, only "certain and sure" local similarities which will contribute to obtain valid results. We think that this choice is logical since, on one hand, it adopts a reasoning based on reliable knowledge to discover similar cases, and on other hand, it do not eliminate cases just because some of their descriptors are not similar to descriptors whose values are uncertain. Finally, skipping uncertain descriptors means considering the case as incomplete and then treating it using the solutions proposed to avoid the problems of knowledge incompleteness.

In case of incomplete knowledge, we propose to proceed first by applying classical techniques of similarity measures. In fact, these techniques ignore the incomplete descriptors when calculating similarity measures. However, restricting these measures to a subset of expected descriptors may probably increase the number of similar cases which are not all necessarily appropriate. In order to address these problems, we propose to apply, as a next step, adaptation measures on the obtained cases to identify the most appropriate among them. To do this, we reuse adaptation measures formulas that were originally proposed by Haouchine et al. to refine the retrieval results regardless of the knowledge imperfections [17]. This refinement is an increasing need, especially, in the case of knowledge incompleteness where a large number of results can be obtained. The idea is to compute the global similarity using weighted local similarities which are measured according to the weight of each descriptor. Ultimately, the descriptors won't be treated the same. Their weights are chosen according to their importance and their impact when making the decision to identify a given case as an appropriate one.

# 6. CASE STUDY AND RESULTS

In this section, we illustrate our solution with a concrete industrial diagnosis example drawn from a previous work presented by Haouchine [18].

The aim is to diagnose the technical breakdowns of a combustion engine which has specific characteristics. We especially note that the engine temperature definition domain ranges between 0°C and 100°C. The combustion engine is operating normally when the temperature reaches 80°C. It can also operate at a temperature of 60°C. However this operating mode is not recommended. Furthermore, we note that in the cooling circuit, the engine temperature must not exceed the value 100°C. As shown in "Figure 2" , we considered three source cases and one target case including an imprecise temperature measure (ds3 descriptor) and also uncertain and incomplete knowledge. On one hand, the value assigned to the descriptor ds9 (noice presence in the turbo compressor with abnormal mode) is considered as uncertain. On the other hand, the values assigned to the descriptors ds6 are incomplete.

| | Index | Target | Source1 | Source2 | Source3 |
|---|---|---|---|---|---|
| ds1 | Engine state | works | works | works | works |
| ds2 | spark plugs | Bad | | Bad | Bad |
| ds3 | Tem° C | 95 | | 100 | 95 |
| ds4 | Sub-zone | | | | |
| ds5 | Injection | Inject. pump | Inject. pump | Inject. pump | |
| | State | Trained | Trained | Trained | |
| | operation | Normal | Abnormal | Normal | |
| ds6 | Filtering | | Monolith | | |
| | State | | Insufficient air | | |
| | operation | | Normal | | |
| ds7 | Explosion | spark plugs | spark plugs | spark plugs | |
| | State | Spark | Spark | Spark | |
| | operation | Normal | Normal | Normal | |
| ds9 | Pressure | Turbo comp. | | Comp. | Turbo comp. |
| | State | Noise presence | | Gaz circulating | Noise presence |
| | operation | A | | N | N |
| ds11 | Friction Movt | Timing belt | Arbres à cannes | | |
| | State | tight | Discont. Movt | | |
| | operation | Normal | Abnormal | | |

**Fig 2: Knowledge extracts from the treated source and target cases .**

In such a case, the most appropriate source case is logically the one whose failing component corresponds to that of the target case. First, we corrected the imprecision marked on the descriptor ds3 corresponding to the temperature of the combustion engine. For this end, we have used fuzzy logic. According to the engine characteristics and as illustrated in "Figure 3", we represent its temperature considering two main ranges of fuzzy values or two fuzzy subsets A1 ([60.79]) and A2 ([81.100]) of the original set E ([0.100]). These two parts are, in fact, the range of variation of the imprecise values which can be approximated using the fuzzy logic and the concept of membership degree. Within this scope, we corrected the imprecise temperature according to the class of triangular fuzzy numbers that we defined using the membership function μ (x):

$$\begin{cases} \mu(x) = 1 \text{ for } x = 80 \\ \mu(x) = 0 \text{ if } |x - 80| \geq 20 \\ \text{and } 0 < \mu(x) < 1 \text{ if } |x - 80| < 20. \end{cases}$$

Considering the degree of membership represented by the graph shown in "Figure 3", the value 95°C belongs to the range A2 (fuzzy set) and will get a membership degree of 0.25 (calculated graphically). It is thereby considered 25% close to 80°C (a value considered far from being close to the terminal 80 of the fuzzy set A2). The imprecise temperature value should then be approximated to 100 ° C.

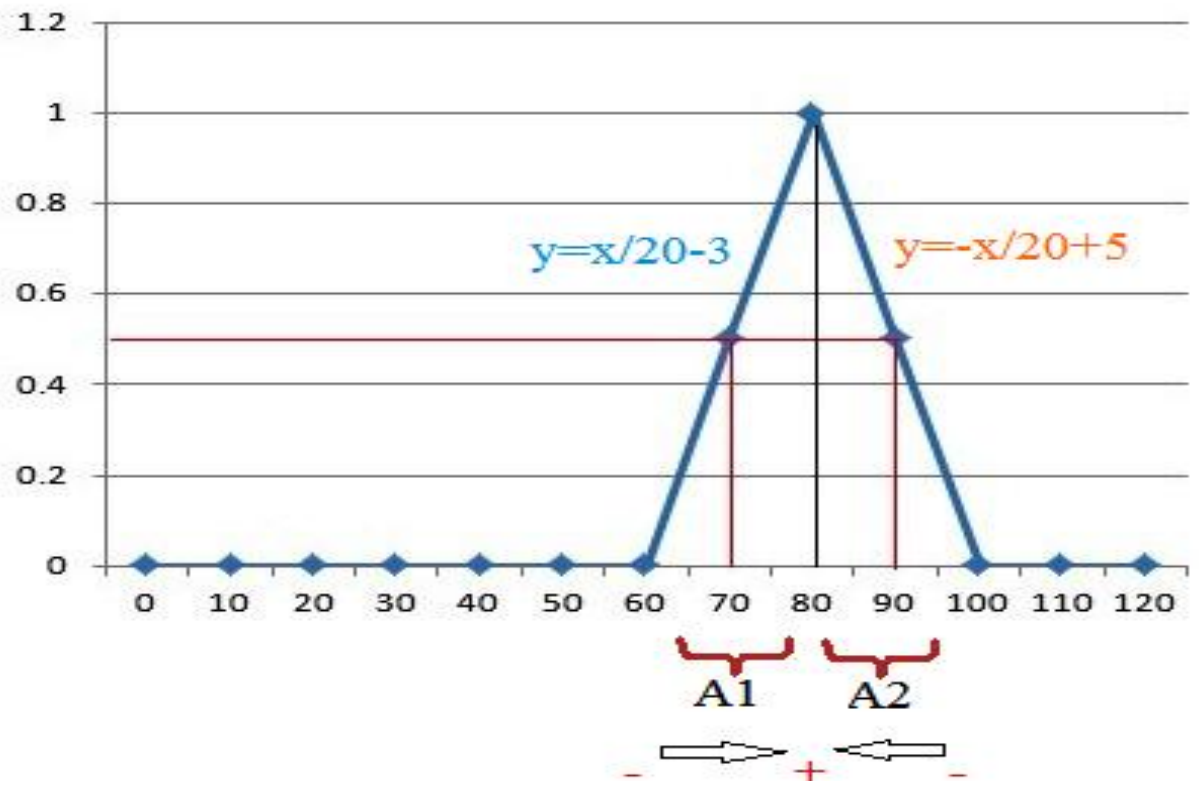

**Fig 3: Graph illustrating the degree of membership to the fuzzy sets A1 and A2.**

To resolve the uncertainty and incompleteness problems, we computed the retrieval measure $M_R$ [17] without taking into account the uncertain descriptor, and then apply an adaptation measure $M_A$ on the retrieved source cases to select appropriately the case that better fits with our target case.

As shown in the following formula (1), the retrieval measure $M_R$ represents an aggregation of four local similarity measures and outcomes in a mathematical formula where m is the number of the problem descriptors. The four local similarity measures are namely: a measure that depends on the membership class of the descriptors values $\varphi^{Value}$ (it is based on the values of the hierarchical levels of the components taxonomy model), a measure which depends on the descriptor state $\varphi^{State}$, a measure that depends on the presence of the descriptors' values $\varphi^{Presence}$ and a measure that depends on the mode of operation $\varphi^{O.M}$.

$$M_R = \frac{\sum_{i=1}^{m} \varphi^{Value} \times \varphi^{State} \times \varphi^{Presence} \times \varphi^{O.M}}{\sum_{i=1}^{m} \varphi^{Presence}} \quad (1)$$

As for the adaptation measure $M_A$ computed through the formula shown as follow in (2), it takes into account the mode of operation of the components giving importance to abnormal operating condition. The operating mode is considered as a crucial criterion in determining the failing component. In this context, we assign to the descriptors of the source cases (dsi) and the target case (dti) a given weight noted λi which is defined depending on the mode of operation. In our case study, we considered the given values:

- If O.M ($ds_i$, $dt_i$) = {normal / normal}→ $\lambda i = 2^0$
- If O.M ($ds_i$, $dt_i$) = {abnormal / normal or normal / abnormal } → $\lambda i = 2^1$
- If O.M ($ds_i$, $dt_i$) = {abnormal / abnormal} → $\lambda i = 2^2$

The descriptors that are in failure mode have high weight because they represent the components that are most likely to be delinquent in the solution space. Besides, the presence of the descriptor value is obviously taken into account. The closeness of the target descriptor value to the class of the source case is also considered when computing the adaptation measure.

$$M_A = \frac{\sum_{i=1}^{m} \lambda i \times \varphi_i^{Presence} \times \varphi_i^{Value}}{\sum_{i=1}^{m} \varphi^{Presence}} \quad (2)$$

The "Table 1" presented below compares the retrieval results obtained by the application of our solution and those obtained using usual and typical approaches without taking into account the knowledge imperfections. It shows that both typical approach and the first phase of our approach dealing with the imprecision and uncertainties identified the source case 2 as the most similar case (0.8 as similarity value obtained using typical approach, and 0.96 as similarity value obtained using the first phase of our approach). Finally, applying the final phase of adaptation, as proposed in our approach, has identified the source case 3 as the case which is most appropriate ( 2 as adaptability value).

**Table 1. Our case study results**

| | | Target/ Source1 | Target / Source2 | Target / Source3 |
|---|---|---|---|---|
| **Typical approaches** | $M_R$ | 0.5 | 0.8 | 0.75 |
| **Our approach** | $M_R$ | 0.667 | 0.96 | 0.5 |
| | $M_A$ | 1.5 | 1.73 | 2 |

## 7. CONCLUSION AND PERSPECTIVES

The solution proposed in this paper improves the CBR cycle results by incorporating new techniques in the retrieval phase. Finally, its application to industrial diagnosis has proven that the adopted techniques and measures are adequate for solving problems of imperfect knowledge. These techniques and measures are founded on fuzzy logic and the involvement of weighted descriptors in the calculation of measures. However, to prove that our solution remains relevant regardless of the domain where it is applied in, we are planning, as perspective, to apply our solution to another application domain, with different types of descriptors. An update strategy is also envisaged in order to evolve the descriptors' weights of the cases saved in the case base.

## 8. ACKNOWLEDGMENTS

Our thanks to Professors Bilal HUSSEIN and Yahia RABIH who have contributed in this work.

## 9. REFERENCES

[1] Y. Peng and J.A. Reggia, "Abductive inference models for diagnostic problem solving." *Symbolic Computation*, Springer-Verlag New York, Inc, 1990.

[2] A. Aamodt and E. Plaza, "Case-Based Reasoning: Foundational Issues, Methodological Variations, and System Approaches". *AI Communications*, 7(1), 39-59, 1994.

[3] I. Watson and S. Abdullah, "Developing case-based reasoning systems: A case study in diagnosing building defects", *IEEE Colloquium on Case-Based Reasoning:*

*Prospects for Applications*, N° 1994/057, Digest. March 1994.

[4] G. Zwingelstein "Diagnostic des défaillances : Théorie et pratique pour les systèmes industriels", *Hermès*, 1995.

[5] P.W. Grant, P.M. Harris and L.G. Moseley, "Fault Diagnosis for Industrial Printers Using Case-Based Reasoning". *Engineering Applications of Artificial Intelligence*. 9(2), 163-173, 1996.

[6] Mille, A., Fuchs, B. et Herbeaux, O. " A unifying framework for adaptation in case-based reasoning", Workshop on Adaptation in Case-Based reasoning, European Conference on Artificial Intelligence, ECAI-96, Budapest, Hungary. (1996)

[7] D. Leake, and D. Wilson, "Categorizing case-base maintenance: Dimensions and directions", *Lecture Notes in Computer Science, vol.1488, Springer-Verlag, Berlin*. 1998.

[8] A. Varma "ICARUS: Design and Deployment of a Case-Based Reasoning System for Locomotive Diagnostics. " In *3rd international conference on case-based reasoning (ICCBR-99)*, vol. 1650, pp. 581-595.1999

[9] L.J. Candia, "Gestion des connaissances imparfaites dans les organisations industrielles : cas d'une industrie manufacturière en Amérique Latine", Thèse de doctorat, Institut National Polytechnique de Toulouse, 2001.

[10] I. Grosclaude " Diagnostic abductif temporel - scénarios de pannes", modèles causaux et traitement de l'information. *Thèse de Doctorat*. Université de Rennes I. 2001.

[11] Afnor. Maintenance terminology. *European standard*, NF EN 13306, 2001.

[12] R. Bergmann, K.D. Althoff, S. Breen, M. Göker, M. Manago and S. Wess. "Developing Industrial Case Based Reasoning Applications: The INRECA Methodology", *Lecture Notes in Artificial Intelligence*, LNAI 1612, Springer Verlag, Berlin, 2003.

[13] B. Bouchon-Meunier and M. Christophe, "Logique floue, principes, aide à la décision". Traité IC2, *Série informatique et systèmes d'information*, Lavoisier, 2003.

[14] Cheetham, W., Tenth Anniversary of Plastics Color Matching, Artificial Intelligence Magazine, Volume 26, No. 3, (2005). pp 51 – 61.

[15] A. Cordier, and B. Fuchs, "Apprendre à mieux adapter en raisonnement à partir de cas", 2006.

[16] A. Mille, "Tutoriel: raisonner à partir de cas: principe, théorisation et ingénierie de la connaissance associée", *14e Atelier du Raisonnement à Partir de Cas*, Besançon, France, March 2006.

[17] K. Haouchine, B. Chebel-Morello and N. Zerhouni, "Conception d'un Système de Diagnostic Industriel par Raisonnement à Partir de Cas", *17ème séminaire de raisonnement à partir des cas,* 115-128, Paris, June 2009.

[18] K. Haouchine, PhD thesis :"Remémoration guide par l'adaptation et maintenance de systèmes de diagnostic industriel par l'approche du raisonnement à partir de cas", 2010.

[19] I. El Bitar, Master Thesis: "CBR: design, implementation and improvement of similarity measures applied to the field of industrial diagnosis", Lebanese University, Doctoral School of Sciences and Technology, 2010, unpublished.

[20] I. El Bitar, B. Hussein, F.Z. Belouadha, O. Roudies: "Solutions aux imperfections de connaissances dans le RàPC", 3ème édition des Journées Doctorales en Technologies de l'Information et de la Communication, 2011.